\documentclass[letterpaper, 10 pt, conference]{ieeeconf}  

\IEEEoverridecommandlockouts                              

\usepackage{graphicx} 
\usepackage{amssymb}  
\usepackage{amsmath} 

\title{\LARGE \bf
Efficient Optimization of a Permanent Magnet Array for a Stable 2D Trap*
}

\author{Ann-Sophia Müller$^{1,2}$, Moonkwang Jeong$^{1}$, Jiyuan Tian$^{1}$, Meng Zhang$^{1}$ and Tian Qiu$^{1,3}$
\thanks{*This work was partially funded by the European Union (ERC, VIBEBOT, 101041975) and the German Cancer Research Center (DKFZ).}
\thanks{$^{1}$A-S. Müller, M. Jeong, J. Tian, M. Zhang, and T. Qiu are with Division of Smart Technologies for Tumor Therapy, 
        German Cancer Research Center (DKFZ) Site Dresden, Blasewitzer Str. 80, 01307 Dresden, Germany}%
\thanks{$^{2}$A-S. Müller is also with Faculty of Computer Science, Dresden University of Technology, 01187 Dresden, Germany}%
\thanks{$^{3}$T. Qiu is with Faculty of Medicine Carl Gustav Carus, Dresden University of Technology, 01307 Dresden, Germany and Faculty of Electrical and Computer Engineering, Dresden University of Technology, 01187 Dresden, Germany
        {\tt\small tian.qiu@dkfz.de}}%
}

\begin{document}

\maketitle
\thispagestyle{empty}
\pagestyle{empty}

\begin{abstract}

Untethered magnetic manipulation of biomedical millirobots has a high potential for minimally invasive surgical applications. However, it is still challenging to exert high actuation forces on the small robots over a large distance. Permanent magnets offer stronger magnetic torques and forces than electromagnetic coils, however, feedback control is more difficult. As proven by Earnshaw's theorem, it is not possible to achieve a stable magnetic trap in 3D by static permanent magnets. Here, we report a stable 2D magnetic force trap by an array of permanent magnets to control a millirobot. The trap is located in an open space with a tunable distance to the magnet array in the range of 20 - 120mm, which is relevant to human anatomical scales. The design is achieved by a novel GPU-accelerated optimization algorithm that uses mean squared error (MSE) and Adam optimizer to efficiently compute the optimal angles for any number of magnets in the array. The algorithm is verified using numerical simulation and physical experiments with an array of two magnets. A millirobot is successfully trapped and controlled to follow a complex trajectory. The algorithm demonstrates high scalability by optimizing the angles for 100 magnets in under three seconds. Moreover, the optimization workflow can be adapted to optimize a permanent magnet array to achieve the desired force vector fields.

\end{abstract}

\section{INTRODUCTION}
Millirobots have become highly promising for biomedical applications, such as targeted drug delivery, \textit{in vivo} sensing, and minimally invasive surgery \cite{palagi2018bioinspired}. One of the key challenges however remains in finding an effective actuation method that enables high forces and precise control of the small devices. Among the various actuation strategies, magnetic actuation has shown to be the most promising one, as it offers relatively large forces and torques, has large penetration depth, and is not harmful to the human body \cite{abbott2020magnetic}. Many magnetic actuation methods have been reported, for example, using homogeneous rotating magnetic fields to drive the rotation of helical robots \cite{zhang2009artificial}\cite{ghosh2009controlled}, using magnetic field gradients to actuate micro-particles \cite{kummer2010octomag}\cite{diller2014six}\cite{qiu2014active}, and using the combination of both modalities to drive miniature robots \cite{hu2018small}\cite{kim2022telerobotic}\cite{jeong2024convoy}. 

Permanent magnetic actuation methods, in particular, offer advantages over electromagnetic coils, as they exert higher magnetic forces and torques over larger distances and there is no need for bulky electrical amplifier or coil cooling systems \cite{qiu2022magnetic}. However, the precise control of a millirobot with permanent magnets poses extra challenges. Many previous efforts have been made to tackle these challenges. For example, Shapiro \textit{et al.} showed an array of two magnets that can push therapeutic nanoparticles \cite{shapiro2010two}\cite{sarwar2012optimal}. Ryan \textit{et al.} reported five degrees-of-freedom control using an array of eight permanent magnets to independently control the field strength and the gradient \cite{ryan2016five}. Baun \textit{et al.} used the superimposed field of a homogeneous and a quadrupolar field to control the force generated on superparamagnetic particles \cite{baun2017permanent}. Son \textit{et al.} developed a setup of 100 magnets to form a cylindrical array of 6 cm in diameter around a trapping point to control a millirobot in brain tissue \cite{son2021permanent}. Yet, to the best of our knowledge, it has not been shown that a stable 2D magnetic force trap can be formed in an open space, where the permanent magnet array is only placed on one side of the trapping point with a considerably useful distance (cf. Fig. 1).
\begin{figure}[ht]
      \centering
\includegraphics[width = 7.5cm]{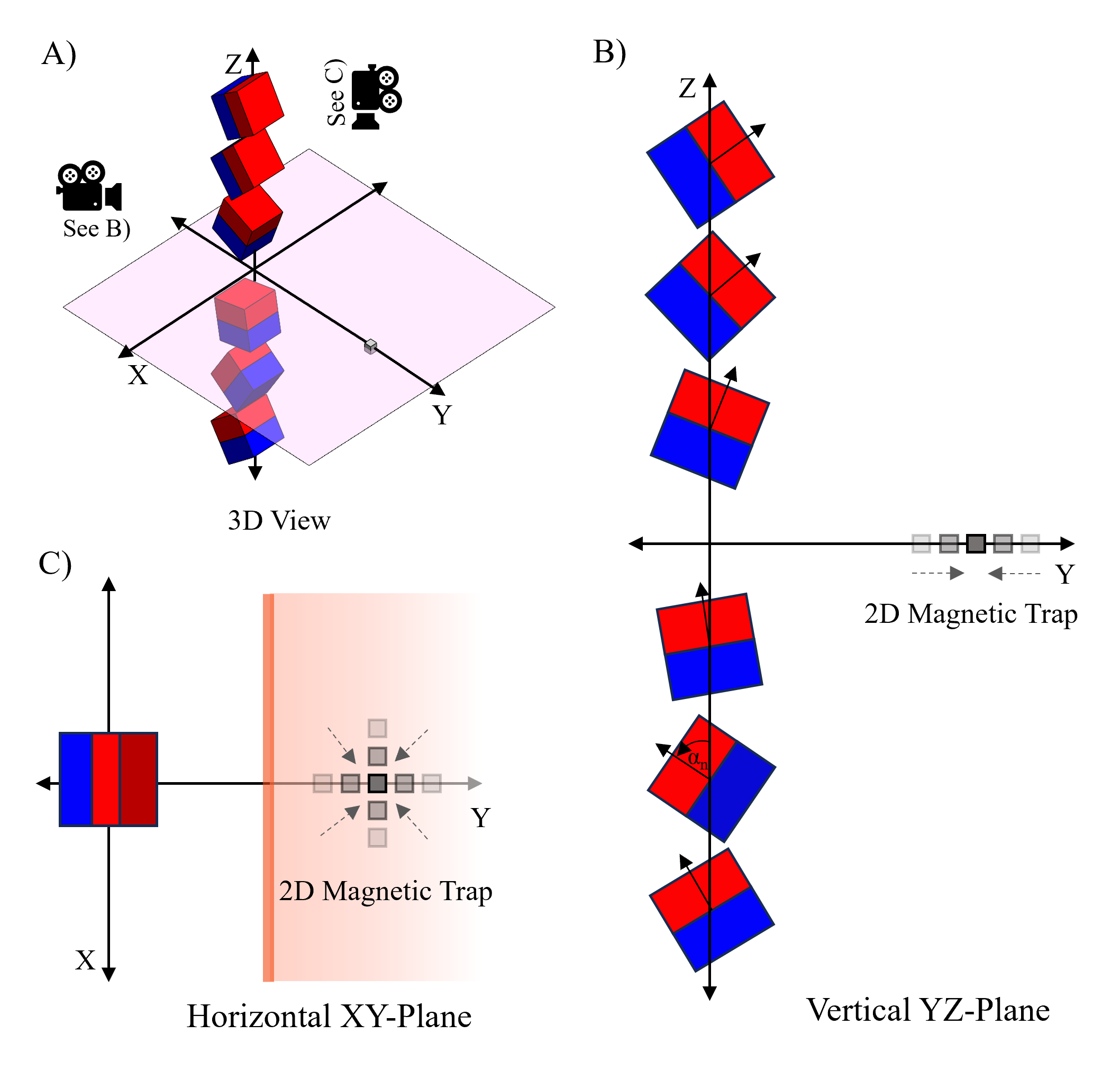}
\caption{Schematic of the stable 2D trap generated in XY-plane, induced by the magnetic force vector field of permanent magnets positioned on the Z-axis. The magnet array located at a far distance and only on one side of the trapping position, making it suitable for biomedical applications.}
      \label{fig1}
   \end{figure}
   
Using the magnetic dipole model \cite{abbott2020magnetic}, second-order nonlinear solvers (e.g., Newton-based methods) can be used to optimize a magnetic array to achieve a desired force vector field, here also referred to as force field. However, computing the Hessian is often costly, making these solvers inefficient for large parameter spaces \cite{nocedal1999numerical}.

Recently, neural networks (NNs) have been reported to prove useful for solving such inverse problems. Pollok \textit{et al.} have shown how deep learning can be used to design structures of permanent magnets given a desired magnetic field \cite{pollok2021inverse}. Tewari \textit{et al.} reported how to use deep NNs to optimize a multi-element surface magnet, based on given field properties \cite{tewari2022deep}. These solutions however rely on image representations of the design space and complex convolutional encoder-decoder architectures.

Here, for the first time, we show the possibility of forming a stable 2D magnetic force trap in open space with only two magnets placed to the side of the trapping point. We further report a novel GPU-accelerated optimization framework to design a magnetic array to form such a trap, given the desired trap distance, magnet properties, and amount of magnets. We demonstrate that it is possible to find an optimized solution by solely using a first-order optimizer - Adam \cite{kingma2014adam}, without any NN layers or second-order gradient computations. The algorithm is implemented using PyTorch \cite{paszke2019pytorch} (version 2.3.0 compiled with cuda version 11.8) allowing for rapid optimization by leveraging its automatic differentiation and GPU acceleration capabilities. PyTorch \cite{paszke2019pytorch} furthermore allows for flexible integration of more sophisticated machine learning techniques for physics-informed optimization. 

We evaluate the quality of our tensor implementation of the magnetic dipole model \cite{abbott2020magnetic} by comparing the simulated magnetic field of an optimized solution with both, a simulation based on commercial finite element software and measurements on a physical prototype. Finally, we show that by using only two magnets, a millirobot is stably trapped in a 2D large open space with a radius of 25 mm at a distance of 89 mm from the magnet array. The millirobot's position can be precisely controlled, enabling it to follow a desired in-plane trajectory. The setup balances simplicity and effectiveness. The magnet placement optimizes the field gradient at the trap location for stable confinement with adequate distance at a realistic human interaction range.

\section{MATERIALS AND METHODS}

\subsection{Optimization algorithm of the magnet array}
The objective is to design a magnetic gradient force vector field that consistently drives a magnetic object back toward the center of a defined area in XY-plane (see Fig. \ref{fig1}C). The center is mandated by the desired position of the trap here defined as $\mathbf{P}_{trap}(y_{trap}) = (0, y_{trap},0) \in \mathbb{R}^3$. The subspace $S\subset \mathbb{R}^3$, in which the force will be evaluated, can be defined as a discrete grid with $i \times j \times 1$ points where each point $\mathbf{P}_{grid_{ij}}(x_{i},y_{j}) = (x_{i},y_{j},0) \in S$ in the grid is evenly spaced around the trapping point within given area bounds. As default we chose $i=j=20$, $y_{trap} = 89\text{ mm}$ and we chose to evaluate the force in a $20 \text{ mm}\times20\text{ mm}$ square area around that point. 
   \begin{figure}[ht]
      \centering
\includegraphics[width=7.5cm]{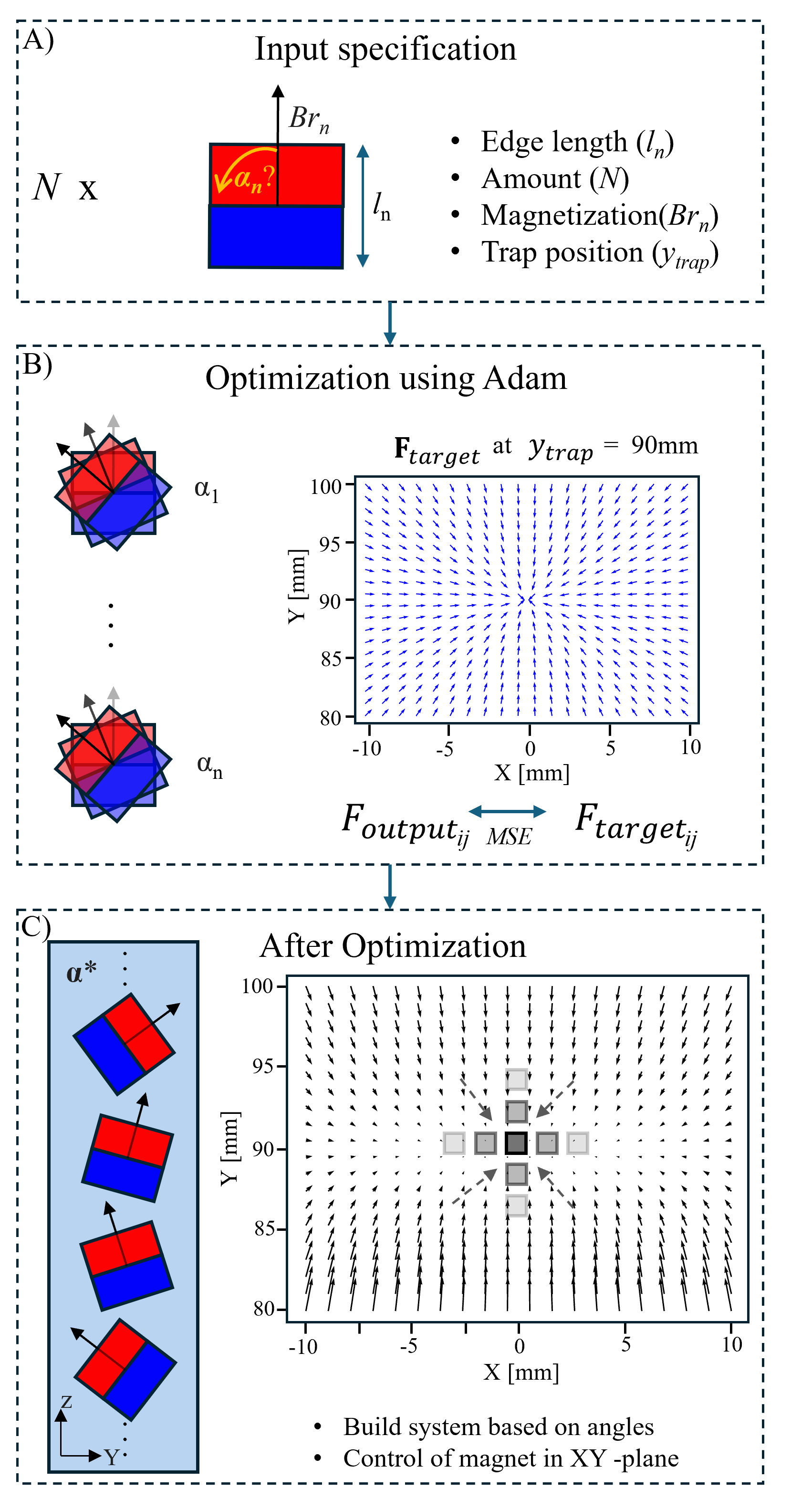}    
      \caption{Optimization workflow. A) The input parameters, such as the magnets' edge length, the magnetization, and the desired distance of the magnetic trap are defined. B) The target force field is generated based on the given trap position, and the optimization algorithm iteratively identifies the optimal angles of the magnets in the array to minimize the mean squared error (MSE) between the resulting and target force fields. C) The physical setup is built based on the optimization results to trap and control a magnetic object in the XY-plane. }
      \label{fig2}
   \end{figure}

Given an even amount of cubic magnets in the array, the positions of the magnets in the array denoted $\mathbf{P}_{mag_n}(z_{n}) = (0,0,z_{n}) \in \mathbb{R}^3$ (where $mag_n$  denotes the $n^{th}$ magnet) are calculated to be symmetrically distributed above and below the center of the Z-axis (cf. Fig. \ref{fig1}B). The spacing between the magnets is, by default, determined based on their face diagonal size ($d_{face} = \sqrt{2}l_{n}+s$) to prevent interference between adjacent magnets where $s$ determines optimal extra space between the magnets an $l_{n}$ denotes the magnets' edge length. After initialization, the positions of the magnets stay fixed with the only variable degree of freedom being the rotational angle around the X-axis $\alpha \in [0,360] $, which is adjusted during the optimization procedure. In our setup, $\alpha$ describes counterclockwise rotations starting from the positive Z-axis (cf. Fig. \ref{fig2}A).

Since, for practical applications, we consider that the distance to the trap should be at least 1.5 times the diagonal length of each permanent magnet \cite{petruska2012optimal}, the magnetic dipole model \cite{abbott2020magnetic} can be used to estimate the force exerted by the array on another magnetic source with an accuracy over 99\% \cite{son2021permanent}. Therefore, given our subspace $S$ we can obtain the force vector field $V$ by the transformation $\mathbf{F}_{ij}:S \rightarrow V, \mathbb{R}^3 \rightarrow \mathbb{R}^3 $ where 
\begin{align}
\mathbf{F}_{ij}(x_i,y_j)=\sum_{n\in N}^{}\mathbf{f}_{nij}
(x_i,y_j,\mathbf{P}_{mag_n},\alpha_{n}) \label{eq:forcefield}
\end{align} is a vector in $V$ representing the force at point $(x_i,y_j)\in S$. 
\begin{align}
\mathbf{f}_{nij} = &\ \frac{3\mu_0}{4\pi\parallel \mathbf{r}_{nij}\parallel^3}
((\hat{\mathbf{r}}_{nij}^T\mathbf{m}_{ij})\mathbf{m}_n + (\hat{\mathbf{r}}_{nij}^T\mathbf{m}_n)\mathbf{m}_{ij}  \nonumber \\
&+ \left( \mathbf{m}_{n}^T\mathbf{m}_{ij} - 5(\hat{\mathbf{r}}_{nij}^T\mathbf{m}_n)(\hat{\mathbf{r}}_{nij}^T\mathbf{m}_{ij}) \right) \hat{\mathbf{r}}_{nij})\label{eq:force}
\end{align}
cf. eq 19 in \cite{abbott2020magnetic}. 

Here, $\mathbf{f}_{nij} \in \mathbb{R}^3$ is the force vector exerted by a single magnet $mag_{n}$ in our magnet array, on a second magnet $mag_{ij}$ (here our millirobot), if placed at grid position $\mathbf{P}_{{grid}_{ij}}$. A bold $\mathbf{m} \in \mathbb{R}^3$ is used to represent the moments of the respective magnets where
\begin{align}
    \mathbf{m}_n  = \begin{pmatrix}
    1 &0 & 0\\
0&\cos{\alpha_{n}} & -\sin{\alpha_{n}} \\
0&\sin\alpha_{n}& \cos{\alpha_{n}} 
\end{pmatrix}
\begin{pmatrix}
    0\\
    0\\
    \frac{Br_{n}}{\mu_{0}}l_{n}^3 
\end{pmatrix}
\end{align}
is the magnetic moment of a single magnet in the magnet array with remanence factor $Br_{n} \in \mathbb{R}$, rotated in our setup by $\alpha_{n}$ around the X-axis. We assume that the direction of the moment of the millirobot will align with the direction of the flux density at the evaluated grid point. Therefore, we consider $\mathbf{m}_{ij} = Br\hat{\mathbf{B}}_{ij}$ where $Br \in \mathbb{R}$ is the remanence of the millirobot and $\mathbf{B}_{ij} = \sum_{n\in N}^{} \mathbf{b}_{nij} \in \mathbb{R}^3$ the magnetic flux density created by the magnetic array \cite{abbott2020magnetic} at $\mathbf{P}_{{grid}_{ij}}$.
\begin{align}
\mathbf{b}_{nij} =\frac{\mu_0}{4\pi\parallel\mathbf{r}_{nij}^3\parallel }(3\hat{\mathbf{r}}_{nij}\hat{\mathbf{r}}_{nij}^T-\mathbf{I})\mathbf{m}_n 
\end{align}
where $\mathbf{b}_{nij} \in \mathbb{R}^3$ is the flux density at $\mathbf{P}_{{grid}_{ij}}$ for a single magnet in the array, and $\mathbf{I}$ is the $3\times3$ identity matrix. $\mu_{0} = 4\pi \times 10 ^{-7} \frac{N}{A^2}$is the vacuum permeability of free space,  $\mathbf{r}_{nij}=\mathbf{P}_{{grid}_{ij}}-\mathbf{P}_{mag_n}$ is the distance vector between the force evaluation points and the magnetic source and $\hat{.}$ denotes unit vectors.

In an ideal force trap, all force vectors within the $i \times j$ grid would point towards $\mathbf{P}_{trap}$ within a given area. It results in
\begin{align}
    \mathbf{F}_{{target}_{ij}}(x_{i},y_{j})=\frac{\mathbf{P}_{{trap}}-\mathbf{P}_{{grid}_{ij}}}{\left\| \mathbf{P}_{{trap}}-\mathbf{P}_{{grid}_{ij}}\right\|} \forall \,\, (x_{i},y_{i})\in S.
\end{align} The resulting unit vectors serve as target force field directions for our optimization problem and are visualized in Fig. \ref{fig2}B.

In each optimization step we calculate the direction of the resulting force vectors as
\begin{align}
    \mathbf{F}_{{output}_{ij}}(x_{i},y_{j}) =  \frac{\mathbf{F}_{ij}}{\left\| \mathbf{F}_{ij}\right\|}, \forall \,\, (x_{i},y_{i})\in S
\end{align}
We try to find the parameters $\mathbf{\alpha}$ that minimize the optimization function:
\begin{align}
    \boldsymbol{\alpha^*} = \underset{\alpha}{argmin}\ \mathcal{L}\left(\boldsymbol{\alpha}\right)  
\end{align}
with
\begin{align}
    \mathcal{L}(\boldsymbol{\alpha}) = \lambda_{1}\mathcal{L}_{1}(\boldsymbol{\alpha}) + \lambda_{2}\mathcal{L}_{2}(\boldsymbol{\alpha})  
\end{align}
where
\begin{align}
    \mathcal{L}_{1}\left(\boldsymbol{\alpha} \right)=\frac{1}{ij}\sum_{ij}\left( \mathbf{F}_{{output}_{ij}}- \mathbf{F}_{{target}_{ij}}\right)^2
\end{align}
and
\begin{align}
    \mathcal{L}_{2}\left( \boldsymbol{\alpha}\right)=(\sum_{ij}\left\|\mathbf{F}_{ij}\right\| - \hat{Y})^2
\end{align}

$\lambda_{1}$ and $\lambda_{2}$ are optional scaling parameters for the respective loss terms. They control the relative importance of each term in the final loss function, allowing to adjust their influence on the overall optimization process. $\hat{Y} \in \mathbb{R}$ represents the desired total force magnitude within the trap. The two loss functions have the following purpose: $\mathcal{L}_{1}$ is the direction loss. It drives the resulting magnetic force vector directions towards the trapping point by penalizing configurations that result in force fields where the direction of the vectors diverges from the target force vector directions. $\mathcal{L}_{2}$ drives the total force magnitude within the trapping area to be close to the desired value $\hat{Y}$. In our case, we want to maximize the force within the trap.

To solve this optimization problem, we use the Adam optimizer \cite{kingma2014adam} with a learning rate of 0.05 and standard beta values of 0.9 and 0.999. While Adam is popular in machine learning, it is a general-purpose optimization algorithm that can be applied to a wide range of gradient-based optimization problems. Besides being computationally efficient and having little memory requirements, the learning rate gets adapted dynamically within the algorithm. This is especially useful for faster convergence, stability, and generalizability as one algorithm is used for different input dimensionalities (i.e. grid sizes, amount of magnets, etc.). Because the positions of the magnets are fixed, and the angle is periodic, an unconstrained optimizer works especially well for our scenario.


Given the algorithm's rapid speed (relying only on first-order derivatives) and to avoid poor local optima due to bad initialization, we initialize  $\boldsymbol{\alpha}$ multiple times with $N$ random values as starting points. The best $\boldsymbol{\alpha}^*$ from these initializations is then returned. By default, we use $ k = 5$ random starts with 300 optimization steps each. To improve efficiency, we introduce a stopping threshold, not performing an additional start if the solution's accuracy exceeds 90\%. Since not all target configurations are physically feasible, the algorithm can lower the threshold (typically by 10\%) if no valid solution is found within $k$ starts. This process repeats for up to $c=3$ iterations by default.

We run the algorithm on a workstation with RTX A6000 Raptor Lake-S GT1 (NVIDIA, USA) GPU and Intel i9-14900k (Intel, USA) CPU.

\subsection{Validation of the magnetic setup by experiments and numerical simulation}
To verify the optimization results, we build a prototype system. The goal is to create a stable trap at 89 mm along the Y-axis from the center of the array of magnets. Four NdFeB N40 permanent magnets (Q-51-51-25-N, Supermagnete, Germany) are assembled to two big cubic magnets ($50.8 \times 50.8 \times 50.8 \text{ mm}^3$), which are positioned with 120 mm distance (center-to-center, Z-direction). Each cube magnet assembly is rotated based on the optimized result, which results in 341\textdegree\ for Mag1 and 19\textdegree\ for Mag2 (c.f. Fig. \ref{fig3}A). 

To align each magnet’s position and angle with the optimization, a magnet housing was designed using SolidWorks (version 2022, Dassault Systemes, France) and 3D-printed (X1 carbon, BAMBULAB, UK) using a thermoplastic filament (PLA basic, BAMBULAB). The assembly of the magnets is placed on the XYZ stage (LTS300C/M, Thorlabs GmbH, Germany) and controlled by a customized code (MATLAB R2023b, MathWorks, USA). This is done to move the magnet array along predefined paths to control the millirobot to follow these movements. 

The resulting magnetic flux density was measured by a magnetometer (Hall Probe C with F3A Magnetic Field Transducers, SENIS, Switzerland). The millirobot was printed using the Up-photo material on a two-photon 3D printer (Nano One, UpNano GmbH, Austria). The overall size is 2.75 mm in length and 1.5 mm in diameter. The tip is designed as a dome shape, and the encapsulated magnet is NdFeB N45 with 2 mm in length and 1 mm in diameter (S-01-01-N, Supermagnete, Germany). The robot was placed at the water-air interface for trajectory control. Additionally, a numerical simulation of the magnetic flux density was performed using the AC/DC Module in COMSOL Multiphysics (v6.2, COMSOL AB, Stockholm, Sweden) to validate the magnetic flux density in different planes. The median remanence value from the data sheet of the magnets \cite{supermagente}, 1.275 T, was used in the simulation.

\section{RESULTS}
\subsection{Qualitative evaluation of the algorithm}
\begin{figure}[ht] 
\centering
\includegraphics[width = 7.5cm]{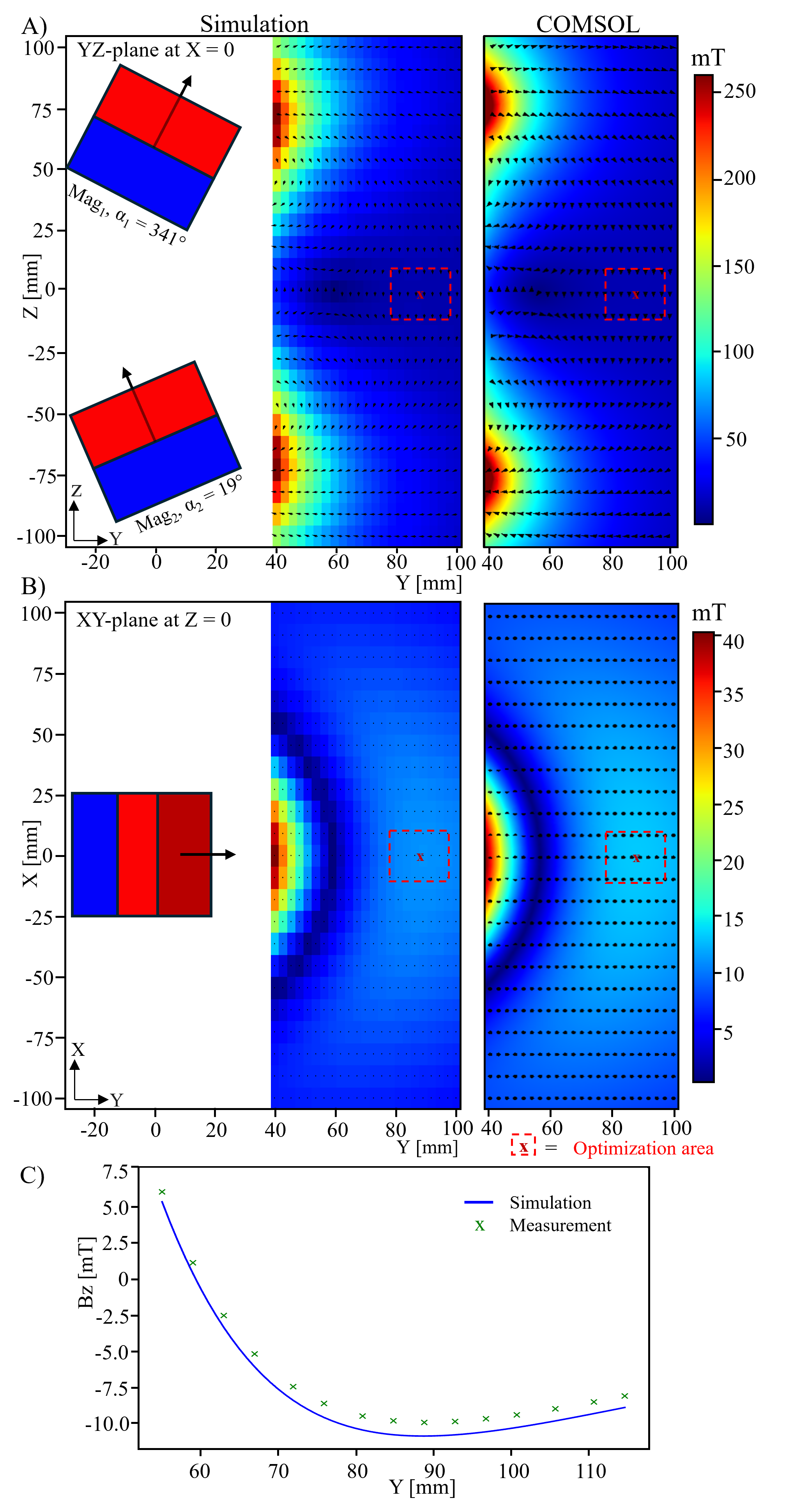}\caption{Qualitative comparison between the developed algorithm and numerical simulation. A) The magnetic flux density vectors in the middle vertical YZ-plane (X = 0). B) The flux density vectors in the middle horizontal XY-plane (Z = 0). The arrows are normalized to represent B-field directions (mainly in the negative Z-direction). The red cross labels the trap position with the surrounding optimization area. C) The simulated $B_z$ component along the center line (X = 0 and Z = 0) compared to the magnetometer measurements.
} \label{fig3}
\end{figure}

Fig. \ref{fig3}A and B show the magnetic flux density in YZ- and XY-plane, respectively. At Z = 0, the $B_z$ changes the sign around 60 mm, corresponding to the highest gradient and maximum pushing force on a magnetic object. Near the trapping position at $\mathbf{P}_{trap}=(0,89\text{ mm}, 0)$, $B_z$ uniformly points downward while $B_x$ and $B_y$ are close to zero. The results of the numerical simulation show good agreement with the results of our algorithm. Additionally, $B_z$ along the Y-axis at Z = 0 and X = 0 is measured and compared to validate the simulation as shown in Fig. \ref{fig3}C. Although we aimed to position the magnetometer probe precisely, minor positioning errors in space may still occur, contributing to deviations.

\subsection{Magnetic force field for a stable 2D trap}\label{sec:3_1}
\begin{figure}[ht]
      \centering
\includegraphics[width =8.7cm]{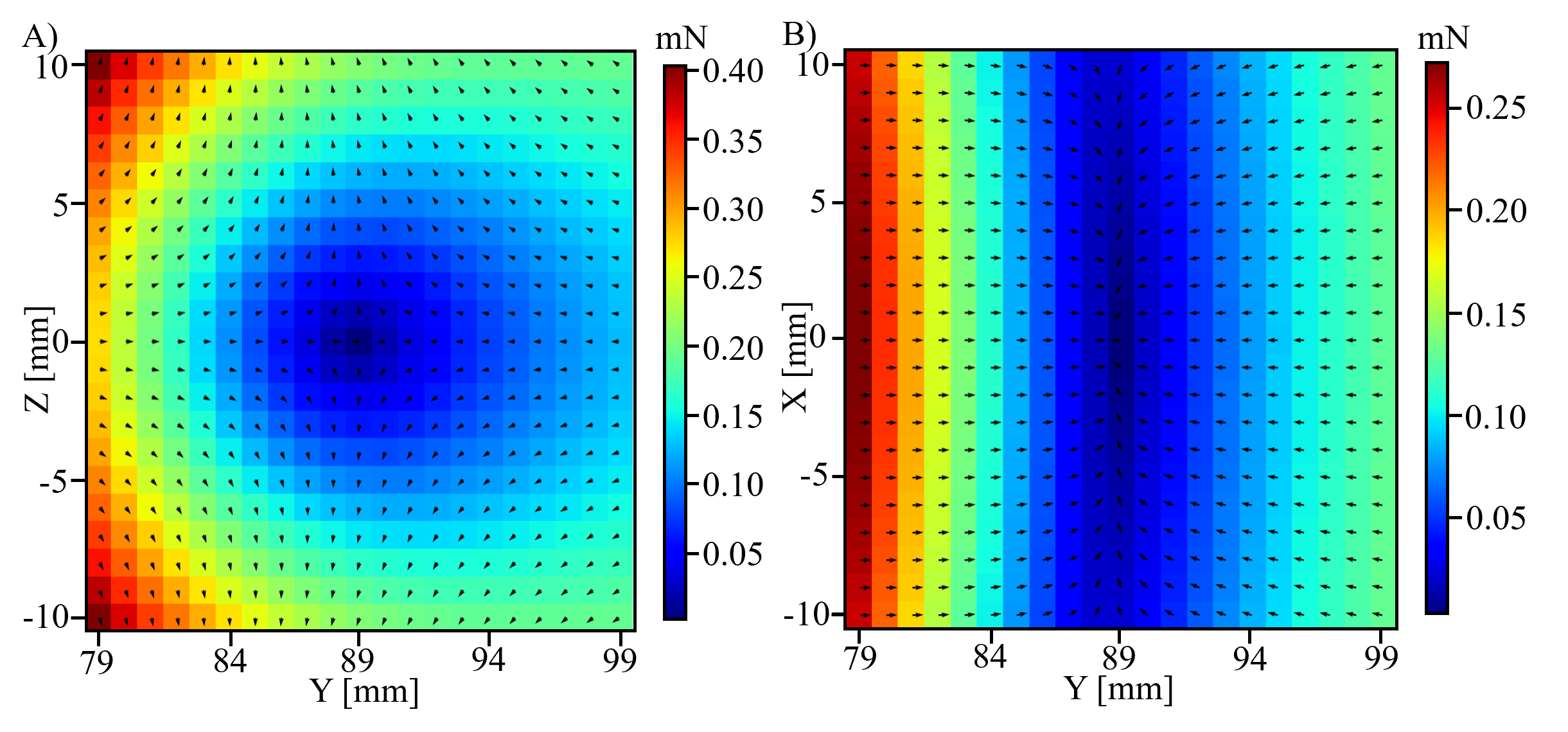}\caption{Magnetic force field analysis. The force vectors in the $20\text{ mm}\times 20\text{ mm}$ optimized region, displayed in A) the YZ-plane X = 0, and B) the XY-plane Z = 0. The arrows represent $\mathbf{F}_{{output}_{ij}}$ after optimization.}\label{fig4}
\end{figure}

Fig. \ref{fig4} shows the magnitude and the direction of the simulated force field in the optimized trapping region of $20\text{ mm} \times 20\text{ mm}$ on the middle vertical YZ-plane ($X = 0$) and the middle horizontal XY-plane ($Z=0$), respectively. In the YZ-plane, it is obvious that there is no stable 2D trap, only a 1D trap along the central axis where $Z = 0$. However, the vertical forces at our trap position  $\mathbf{P}_{trap}$ are not strong enough to lift our millirobot from the surface against gravity, thus it does not affect the 2D trap in the XY-plane. In Fig. \ref{fig4}B, $\mathbf{F}_{{output}_{ij}}$ is visualized. The trap at the desired location $\mathbf{P}_{trap}$ is visible at the center of the image, with forces near the trap’s center approaching zero. Weaker forces are exerted along the X-axis (Y-axis of the plot).
\subsection{Magnetic force trap properties}
\begin{figure}[ht]
      \centering
\includegraphics[width =8.5cm]{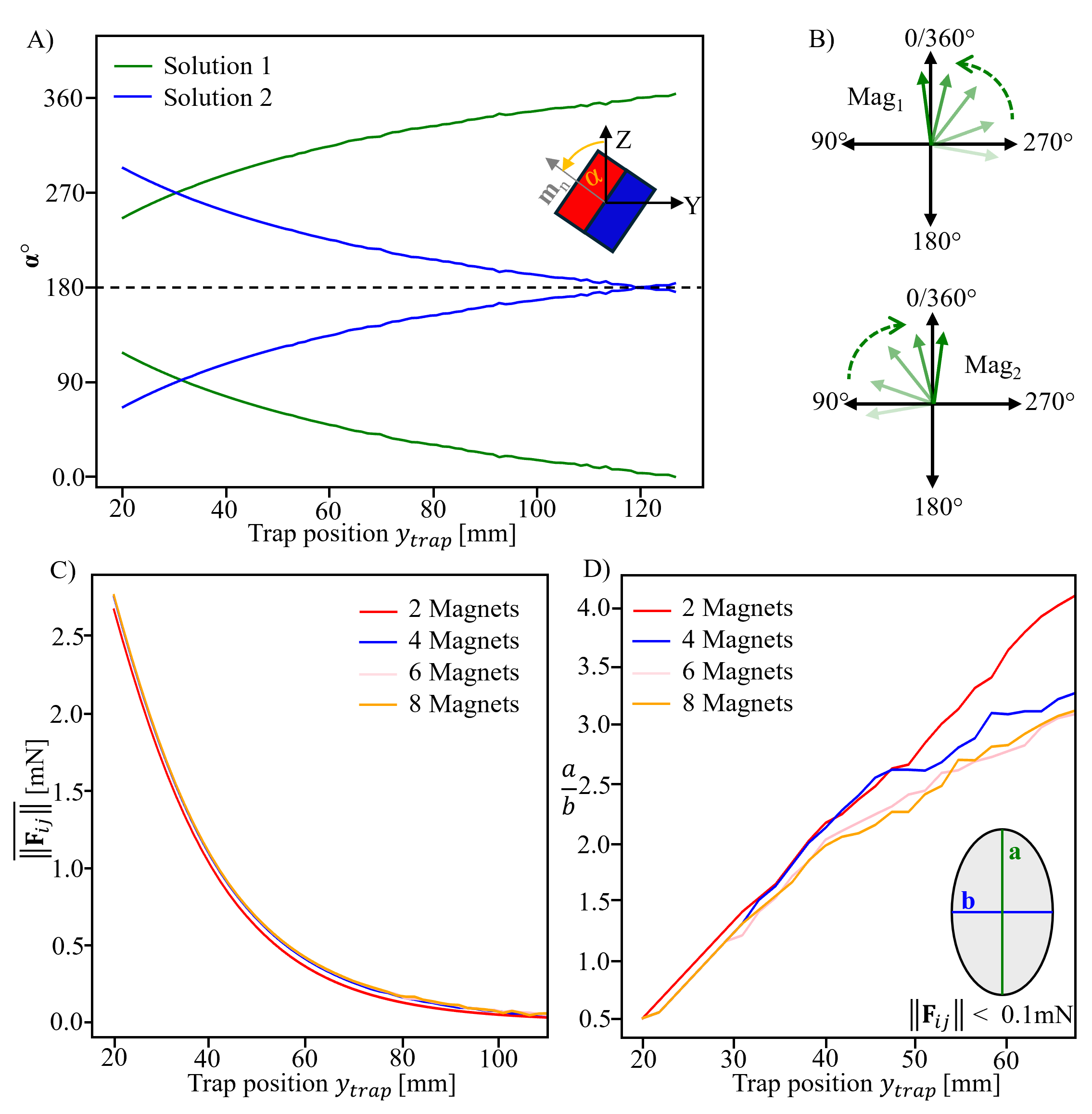}\caption{Properties of the magnetic trapping system. A) Angle - trapping distance configurations. The lines represent the $\alpha$ angles of two magnets as a function of the resulting trapping position. Two solutions can be found to realize one desired trap position. B) Visualization of Solution 1: As the magnetic moments rotate in the direction of the circular arrows, the trap position moves further away. The arrows represent magnetic moment directions. C) Trapping force strength variation. The average force magnitude within a 10 mm radius region is plotted over an increasing trap distance, for different numbers of magnets with an edge length of 50.8 mm. D) Aspect ratio of the trap area as a function of the trap distance, for different numbers of magnets with edge length of 50.8 mm.} \label{fig5}
   \end{figure}

For the two-magnet setup, we optimized the angles for 100 trap positions between 20 and 130 mm. Due to symmetry, two possible solutions exist for the optimized angles, as shown in Fig. \ref{fig5}A. The rotational angle for each magnet is plotted as a function of the desired trap position. Notably, both magnets can be rotated continuously to adjust the trap position further away which could enable systems that create controlled trajectories by adjusting magnet angles. Fig. \ref{fig5}B visualizes these rotations. 
Fig. \ref{fig5}C illustrates how the average force magnitude within a 10 mm radius region, with the trap as the center, decreases with the trap distance from the magnet array. Results are shown for configurations of different numbers of magnets with an edge length of 50.8 mm. Interestingly, adding more magnets does not significantly increase the force, which is likely due to the greater distance between the outer magnets and the trap center, reducing their contribution. However, Fig. \ref{fig5}D suggests that increasing the number of magnets results in a more circular trap, which enhances stability for moving a millirobot. The figure shows the moving-averaged ratio of the of the major to minor axis lengths of the elliptical region where the force magnitude falls below 0.1 mN. A perfect trap would have a ratio of 1, forming a circle. As shown in Fig. \ref{fig4}B, the forces along the perpendicular sides of the magnetic array are weaker than along the parallel sides, leading to an oval-shaped trap. With just two magnets, the major axis already extends to the trap boundary at around 70 mm. Increasing the number of magnets to four results in a more symmetric trap. Notably, compared to the difference between two and four magnets, the differences between four, six, and eight magnets are smaller, with the latter two configurations yielding nearly identical results. This is also likely due to the diminishing contribution of magnets further from the center.
\subsection{Algorithm characteristics}
To evaluate the algorithm's performance, we tested it on an input space of 100 magnets (1 mm edge length) and let it generate traps at 100 positions within a 5-20 mm range. The magnets in the array and at the trap position were defined as N45. We set the algorithm to perform 20 random starts and 300 iterations each, with a 95\% accuracy stopping criterion. Since the ground truth for the maximum total force magnitude in the optimized region is unknown, tuning the value of $\hat{Y}$ is crucial. If $\hat{Y}$ is set too high, the error of $\mathcal{L}_2$ overly influences the optimization, potentially causing the algorithm to miss a trap by focusing excessively on large forces. A stable approach was to first compute a solution with $\lambda_{2} = 0$, then setting $\hat{Y}$ to $\gamma$ times the resulting overall force in the trap, and rerunning the algorithm with $\lambda_{1} = \lambda_{2} = 1$. For this evaluation, we set $\gamma = 1.5$. Despite the challenging task, the algorithm required an average of 7.7 seconds per trap position to complete. However, an optimum was already found after an average of 2.7 seconds. The algorithm achieved an average accuracy of 93\%, with better results closer to the array - a noteworthy outcome, given the increased difficulty of generating traps at greater distances in this setup. Notably, each magnet was assigned a specific angle, demonstrating that the algorithm considers all magnets, not just those near the center. To further assess performance, we calculated the center of the resulting trap, defined as the average location of points with the lowest force within the trapping region. In 89\% of cases, the trap's center aligned with the target, while in 10\% it was shifted by 0.1 mm towards and once, by 0.1 mm away from the array. This suggests the algorithm may sometimes improve force strength by slightly adjusting the trap's position closer to the array while maintaining proximity to $\mathbf{F}_{{target}_{ij}}$. Overall, the algorithm required under 1800 steps to find an optimum for 100 magnets, achieving a 252-order magnitude improvement over the brute force method’s worst-case scenario of $360^{100}$ steps.
\subsection{Physical experiments}
\begin{figure}[h]
      \centering
\includegraphics[width =7.7cm]{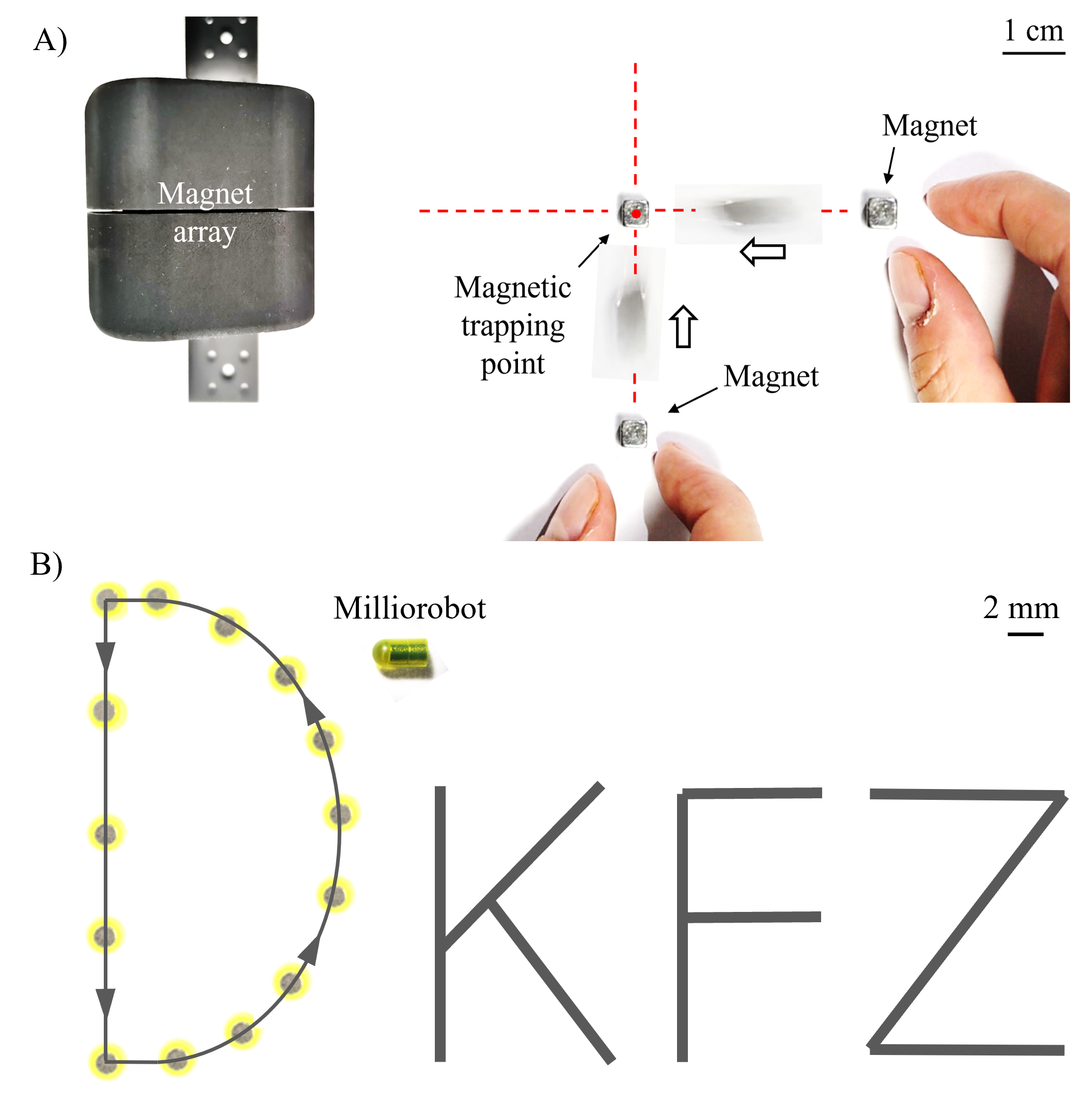}\caption{Experimental verification of algorithmic results. A) Stable trapping. A magnet successfully and quickly returns to the trap position when moved away by hand. B) A cylindrical millirobot is trapped and controlled by moving the magnet array to follow a trajectory of the letter "D" on a water-air interface.} \label{fig6}
\end{figure}

We built a simplified physical prototype using two permanent magnets to verify the calculation. Fig. \ref{fig6}A illustrates how a cubic magnet with a 5 mm side length is quickly drawn back to the trap position, when it is moved away by hand, even beyond the $20 \text{ mm} \times 20\text{ mm}$ (see the supplementary video). Fig. \ref{fig6}B shows the successful control of a millirobot to follow the pre-defined trajectory of the letter "D". The robot follows the movement of the magnetic array with an average accuracy of $\pm$0.1 mm with a mean velocity of 16 mm/s and 0.5 mm/s for the straight and the curve lines, respectively. These results demonstrate the system's capability for wireless trapping and precise manipulation of a millimeter-sized object without feedback control, thanks to the special design of the magnetic trap.

\section{CONCLUSIONS}
In this work, we demonstrated for the first time an efficient method to use an array of permanent magnets to create a stable force trap in the 2D plane at a large distance, enabling precise control of biomedical millirobots. Additionally, we report a GPU-accelerated algorithm based on Adam optimization to design the magnet array to achieve a desired 2D force vector field - here a trap. It may enable versatile machine learning-based algorithms to optimize more degrees of freedom of the magnet array in future works, potentially enabling more complex force or magnetic fields.

\section{ACKNOWLEDGEMENT}
M.J. thanks Institute of Physical Chemistry, University of Stuttgart for the access to computing power and software licenses. 

\addtolength{\textheight}{-12cm}   





\bibliographystyle{ieeetr}  
\bibliography{IEEEfull}  

\end{document}